\documentclass{article} 
\usepackage{iclr2021_conference,times}

\usepackage{amsmath,amsfonts,bm}

\def\eqref#1{equation~\ref{#1}}

\def\1{\bm{1}}

\DeclareMathAlphabet{\mathsfit}{\encodingdefault}{\sfdefault}{m}{sl}
\SetMathAlphabet{\mathsfit}{bold}{\encodingdefault}{\sfdefault}{bx}{n}

\newcommand{\R}{\mathbb{R}}

\usepackage{hyperref}
\usepackage{url}
\usepackage{graphicx}
\usepackage{xcolor}
\usepackage{amssymb}
\usepackage[noend]{algorithmic}
\usepackage{multirow}
\usepackage{pgfplots}
\usepackage{booktabs}

\title{FSPN: A New Class of Probabilistic Graphical Model}

\author{Ziniu Wu$^{\#}$, Rong Zhu$^{\#, *}$, Andreas Pfadler, Yuxing Han, Jiangneng Li, Zhengping Qian, \\  \textbf{Kai Zeng,~Jingren Zhou} \\ \\
	Alibaba Group, Hangzhou Province, China \\
	\{ziniu.wzn, red.zr, andreaswernerrober, yuxing.hyx, jiangneng.ljn, zhengping.qzp, \\ zengkai.zk jingren.zhou\}@alibaba-inc.com \\
	$\#$: Equally Contribution  $*$: Corresponding Author
}

\iclrfinalcopy

\newcommand{\D}{\textsc{d}}
\renewcommand{\H}{\textsc{h}}

\renewcommand{\L}{\textsc{l}}
\newcommand{\N}{\textsc{n}}
\newcommand{\W}{\textsc{w}}
\renewcommand{\R}{\textsc{r}}

\newcommand{\pa}{\text{pa}}

\newcommand\revise[1]{\textcolor{black}{#1}}

\definecolor{mywt}{RGB}{255, 230, 153}
\definecolor{mytp}{RGB}{169, 209, 142}
\definecolor{mywh}{RGB}{239, 182, 176}
\definecolor{mywf}{RGB}{180, 199, 231}

\begin{document}

\maketitle

\vspace{-2em}
\begin{abstract}
We introduce \underline{f}actorize-\underline{s}um-\underline{s}plit-\underline{p}roduct \underline{n}etworks (FSPNs), a new class of probabilistic graphical models (PGMs). \revise{FSPNs are designed to overcome the drawbacks of existing PGMs in terms of estimation accuracy and inference efficiency.} Specifically, Bayesian networks (BNs) have low inference speed and performance of  \revise{tree-structured} sum-product networks(SPNs) significantly degrades in presence of highly correlated variables. FSPNs absorb their advantages by \emph{adaptively} modeling the joint distribution of variables according to their dependence degree, so that one can simultaneously attain the two desirable goals---\emph{high estimation accuracy }and \emph{fast inference speed}. We present efficient probability inference and structure learning algorithms for FSPNs, along with a theoretical analysis and extensive evaluation evidence. \revise{Our experimental results on synthetic and benchmark datasets indicate the superiority of FSPN over other PGMs.}
\end{abstract}

\section{Introduction}

Probabilistic graphical models (PGMs), a rich framework for representing probability distributions, \revise{have commonly been used for} building \revise{\emph{accurate}} and \emph{tractable} models for high-dimensional complex data. In comparison with deep generative models containing many hidden layers, which are ``black-box'' approximators, PGMs \revise{without latent variables} tend to be much more interpretable and faster in inference. They are more appliable in lots of tasks such as online causal inference~\citep{pearl2009causal} and query selectivity estimation~\citep{2001SigmodGreedy}, which have strict requirements on both inference accuracy and speed of the deployed models. \revise{Therefore, PGMs have recently re-attracted considerable research interests in the ML community~\citep{zheng2018dags, scanagatta2019survey, paris2020sum}. Much efforts~\citep{VarLearnSPN, vergari2015simplifying,SPGM,mspn,shao2019conditional,sharir2018sum,choiprobabilistic,2014Cutset,darwiche2009modeling,boutilier2013context} have been devoted to improving the accuracy and tractability (inference speed) of PGMs.}

\textbf{Challenges of PGMs.}
\revise{In  the most well-known class of PGMs}---Bayesian networks (BNs), a set of random variables is modeled as a directed acyclic graph (DAG) where each variable (node) is conditionally independent of others given its parents. BNs can accurately and compactly model data distributions. \revise{However, although sometimes tractable, marginal probability inference on BNs is generally intractable.} Even worse, structure learning of BN is NP-hard~\citep{chickering1996learning}. Another class of PGMs---Markov random fields (MRFs), model the joint \revise{probability density function (PDF)} as an undirected graph. \revise{However, marginal probability inference on MRFs is also difficult~\citep{PGM,murphy2012machine}.} Even computing the likelihood of a single data point is intractable.

To improve tractability,~\citep{SPN} proposed a new class of PGMs, the sum-product networks (SPNs). \revise{ SPNs are recursively defined as weighted sums or products of smaller SPNs on simpler distributions}. \revise{The most widely used SPN has a tree-structure, whose inference time is linear w.r.t.~the number of nodes. They have high expressive efficiency on weakly correlated variables~\citep{SPN_expressive}.} However, for highly correlated variables, their joint PDF is difficult to split into smaller ones where variables are locally independent. \revise{The learned SPN has large size and poor generality.} \revise{Prior works have tried to extend SPNs by incorporating BNs or MRFs into SPNs~\citep{VarLearnSPN,vergari2015simplifying,SPGM} or learning directed acyclic graph (DAG)-structured SPNs~\citep{mspn,dennis2015greedy}. However,  these models either slow down the inference speed or hard to learn.
Some works~\citep{shao2019conditional, sharir2018sum} designs variations of SPNs to model conditional PDFs, but they are not suitable for evidence or marginal probability computation.
}

\revise{
In summary, existing PGMs still have some drawbacks in terms of estimation accuracy or inference speed. Designing highly accurate and tractable PGMs remains a challenging task.
}

\textbf{Our Contributions.}
The key reason of existing PGMs' drawbacks \revise{arise from that} they only utilize a single approach to decompose the joint PDF. \revise{BNs rely on conditional factorization, which is accurate but difficult for inference. Tree-structured SPNs use locally independent factorization, which cannot work well in presence of highly correlated variables.} This naturally leads us to the following question: \textit{if we could combine the strength of the two factorization approaches, is it possible to design a new type of PGM that is both accurate in estimation and fast for inference?}
To this end, we propose \underline{f}actorize-\underline{s}um-\underline{s}plit-\underline{p}roduct \underline{n}etworks (FSPNs), \revise{a versatile PGM aiming at this goal.} 

The main idea of FSPN is to \emph{adaptively} decompose the joint PDF of variables based on their dependence degree. Specifically, 
FSPN separates the set of highly correlated variables from the rest by conditional factorization without loss of precision and processes each part accordingly.
For highly correlated variables, their values are \emph{interdependent}, so a multivariate PDF using some dimension reduction techniques~\citep{regression, wang2015joint} is more suitable to model them.
For the remaining weakly correlated variables, local independence commonly exists, so that their joint PDF can be split into small regions
where they are mutually independent. FSPN recursively applies the above operations to model the joint PDF in a compact tree structure.
We show that, FSPN is a very general PGM, which subsumes \revise{tree-structured SPNs} and discrete BNs as special cases.

FSPNs absorb existing PGMs' advantages \revise{while overcoming} their drawbacks.
\revise{First, expressive efficiency of FSPNs is high, as two factorization approaches are used for variables with different dependence degree in an adaptive manner.}
Second, FSPN models are tractable, as their inference time is near linear w.r.t.~its \revise{number of nodes}. Third, structure learning of FSPNs is efficient. A locally optimal structure can be easily and efficiently obtained. In our evaluation, FSPNs consistently outperform \revise{other models} on datasets with varied variable dependence degree \revise{in terms of estimation accuracy, inference speed, model size and training cost}. On a series of PGM benchmark datasets, FSPNs also achieve \revise{comparable performance w.r.t.~the state-of-the-art models}. In summary, our contributions are as follows:
\vspace{-0.5em}

\hspace{1.5em}$\bullet$
We propose FSPNs, a novel and general \revise{class of PGMs} that simultaneously attain high estimation accuracy and fast inference speed (Section~3). 
\vspace{-0.5em}

\hspace{1.5em}$\bullet$
We design an efficient inference algorithm for FSPNs, which runs in near linear time w.r.t.~its node size (Section~4).
\vspace{-0.5em}

\hspace{1.5em}$\bullet$
We devise an efficient structure learning algorithm for FSPNs, which returns a locally maximum likelihood structure (Section~5). 
\vspace{-0.5em}

\hspace{1.5em}$\bullet$
We conduct extensive experiments to demonstrate the superiority of FSPNs on both synthetic and benchmark datasets (Section~6).


\section{Background and Related Work}

In this section, we briefly review some background knowledge and related work. Let $X = \{X_1,\dots, X_m\}$ be a set of $m$ random variables and $D\in \mathbb{R}^{n \times m}$ be a training data matrix sampled from $\Pr(X)$. A PGM aims at building a compact generative model $\Pr_{\D}(X)$ on $X$ such that: \revise{
1) $\Pr_{\D}(X)$ can be efficiently learned to accurately approximate the joint PDF $\Pr(X)$;
and 2)  marginal probabilities can be inferred efficiently using the model on $\Pr_{\D}(X)$.} Building a PGM that fulfills these two goals remains non-trivial. We review two well-known classes of PGMs as follows.

\textbf{Bayesian Networks (BNs)}
represent the joint PDF $\Pr(X)$ of $X$ as a DAG based on the conditional independence assumption. Such a representation is exact when the BN structure accurately captures the causal relations among variables. Therefore, it is an expressive and explainable PGM. However, BNs have significant drawbacks in terms of inference and structure learning efficiency. First, marginal probability inference for BNs has a high time complexity and is sometimes even intractable. Exact inference methods, such as variable elimination and belief propagation~\citep{PGM},  have exponential time complexity w.r.t.~its node size. Approximate inference methods, such as sampling~\citep{gelfand2000gibbs,andrieu2003introduction} and loopy belief propagation~\citep{murphy2013loopy}, reduce the time cost but sacrifice estimation accuracy. Second, the structure learning problem for BN is NP-hard. Exact methods rely on either combinatorial  search~\citep{chickering1997efficient} in a super-exponential space of all DAGs or numerical optimization ~\citep{zheng2018dags, yu2019dag} with high time complexity on computing the DAG constraints. Some approximate methods speed up the learning process by heuristic rules~\citep{scanagatta2015learning,ramsey2017million}, \revise{which may make the learned BN structure inaccurate for probability estimation.} \revise{Some prior works have focused on learning tractable BNs by compiling into a circuit~\citep{lowd2012learning} or bounding tree width~\citep{scanagatta2016learning}. On a Chow-Liu tree, inference time is $O(nd^{h + 1})$ where $n$, $d$ and $h$ represents the number of nodes, domain size and tree width, respectively. In our evaluation, the inference time on such BNs is still slow even when $h = 1$.}

Unlike BNs, Markov random field (MRFs) model the joint PDF of variables as an undirected graph. The marginal probability inference on MRFs involve computing the normalizing factor of the potential functions, the complexity of which is exponential w.r.t. the tree width of the underlying graph. 
\revise{Therefore, MRFs are not suitable for computing marginal probabilities and are thus mainly used for data generation and pattern recognition~\citep{li2009markov}.}
\revise{Another tractable PGM related to BNs are cutset networks~\citep{2014Cutset}, which can be regarded as an ensemble of tractable BNs.}

\revise{
\textbf{Sum-Product Networks (SPNs) } model the joint PDF by recursively applying two operations, namely sum and product, to 
split the joint PDF into simpler PDFs to capture contextual independence. Specifically, a sum node represents a weighted sum of mixture models as $\Pr_{\D'}(X') = \sum_{j} w_j \Pr_{\D'_j}(X')$ where $D'_j$ and $w_j$ are the data and weight of the $i$-th child.}  A product node partitions variables $X'$ into mutually independent subsets $S_1, S_2, \dots, S_d$ such that $\Pr_{\D'}(X') = \prod_{i} \Pr_{\D'}(S_i)$. A leaf node \revise{commonly} maintains  a univariate distribution $\Pr_{\D'}(X')$ of a single variable $X'$. The marginal probability can be computed by a bottom-up traversal on SPNs, whose time cost is linear w.r.t.~\revise{the number of nodes}.

\revise{	
In the literature, tree-structured SPNs can be efficiently learned by a number of methods~\citep{DisLearnSPN,LearnSPN,GreedyLearnSPN,BayesLearnSPN}. Thus, they are most widely used. However, tree-structured SPNs can not work well in presence of highly correlated variables in $X$. In this situation, a single product node is unable to split these variables and SPN would repeatedly add sum nodes to split $D'$ into very small volumes, i.e., $|D'| \! = \! 1$ in extreme. This large structure has low inference speed and poor generality, which degrades its estimation quality.}

\revise{
In order to overcome the drawbacks of tree-structured SPNs, a number of attempts have been made to incorporate BNs or MRFs into SPNs. \citep{SPGM} designs SPGMs, a hybrid model of BNs and SPNs, where sum and product nodes are applied and their children may be BNs modeling partial variables in $X$.
However, inference time for SPGMs is quadratic w.r.t.~the maximum domain size of variables in $X$ with a large factor. \citep{VarLearnSPN} and~\citep{vergari2015simplifying} proposed SPN-BTB and ID-SPN, which apply BNs and MRFs to model $\Pr_{\D'}(X')$ as multivariate leaf nodes where variables in $X'$ can not be easily modeled by sum and product operations. They are more compact than tree-structured SPNs. However, the embedded BNs or MRFs may slow down the inference process. In  our evaluation, their inference time is longer than tree-structured SPNs.
}

\revise{
More general form of SPN structures are DAGs. DAG-structured SPNs tend to be much more compact and efficient for inference than tree-structured SPNs, as they merge redundant sub-structures into a singleton unit in DAGs. However, learning an optimal DAG-structured SPN is NP-Hard~\citep{mspn}. Existing solutions obtains the DAG structure by greedy search or heuristic rules~\citep{mspn,dennis2015greedy}  over tree-structured SPNs, whose performance gain may be limited. 
}

\revise{
SPNs can be extended to model conditional PDFs $\Pr(Y | X)$. \citep{shao2019conditional} proposed conditional SPNs (CSPNs).
Each leaf node in CSPNs models $\Pr(Y_i | X)$ on a singleton variable $Y_i$. CSPNs are mainly used for point data and cannot be directly used to compute marginal probabilities on $X$. \citep{sharir2018sum} proposed sum-product-quotient networks (SPQNs), where the quotient operation can model the conditional PDF by dividing the PDFs of its children. However, until now, SPQNs are only a theoretical model, and no structure learning methods have been proposed.
}

\revise{
There also exist other types of PGMs, such as probabilistic sentential decision diagrams (PSDDs)~\citep{2014PSDD}. PSDDs can model the PDF in presence of massive, logical constraints but only work on binary variables.
}

\revise{
\textbf{Summary.}
To model a joint PDF, existing PGMs apply two factorization approaches. BNs and some types of SPNs use \emph{conditional factorization} to capture conditional independence, which is accurate but inefficient for inference and structure learning. Tree-structured SPNs use \emph{independent factorization} to capture contextual independence, which enables fast inference but has poor performance in presence of highly correlated variables. None of them can comprehensively fulfill the desired goals of PGMs. To resolve these drawbacks, we combine their advantages to design a new type of PGMs, which attains both estimation accuracy and inference efficiency at the same time.
}



\section{The FSPN Model}

In this section, we propose the \text{\underline{f}actorize-\underline{s}plit-\underline{s}um-\underline{p}roduct} \underline{n}etwork (FSPN), a new class of PGM. 

\textbf{Main Idea.}
FSPN adaptively applies both  conditional and independent factorization approaches to decompose variables $X$ with different dependence degree. Specifically, let $H$ be a set of highly correlated variables in $X$, e.g., whose pairwise correlation values are above certain threshold.
A tree-structured SPN can not accurately and efficiently model $\Pr(X)$ in presence of $H$. Therefore, we first decompose $\Pr(X)$  as \revise{$\Pr(X) = \Pr(W) \cdot \Pr(H | W)$ where $W = X \!\setminus\! H$} by the conditional factorization approach and model each distribution accordingly.

For \revise{$\Pr(H | W)$}, each value $x$ of \revise{$W$} specifies a conditional PDF $\Pr(H | x)$. To compactly model \revise{$\Pr(H | W)$}, we partition \revise{$W$} into multiple ranges $R_1, R_2, \dots, R_t$ such that for any $x, x'$ in the same  $R_i$, $\Pr(H | x) = \Pr(H | x')$ roughly holds. Then, we only need to maintain one PDF $\Pr_{i}(H)$ for each $R_i$. As variables in $H$ are \revise{highly correlated}, we can model $\Pr_{i}(H)$ as a multivariate PDF using sparse distribution modeling~\citep{wang2015joint} and 
 dimension reduction techniques such as PCA/ICA and piece-wise regression~\citep{regression}. For \revise{$\Pr(W)$}, we can recursively separate each set of highly correlated variables from \revise{$W$} using the above method until the remaining variables \revise{$W' \subseteq W$} are weakly correlated. As local \revise{contextual} independence highly likely holds  on \revise{$W'$}, we can apply the independent factorization approach to further decompose \revise{$\Pr(W')$}. 
 Specifically, \revise{$\Pr(W')$} is split into multiple small regions where variables in \revise{$W'$} are locally independent. Then, for each region, we maintain a univariate distribution $\Pr(X_j)$ for each $X_j \in W'$.

\textbf{Formal Definition.}
Given a set of variables $X$ and training data matrix $D$, let $\mathcal{F}$ denote the FSPN modeling the joint PDF $\Pr_{\D}(X)$.
Each node $N$ in $\mathcal{F}$ either represents a PDF $\Pr_{\D'}(S)$ of a variable subset $S \subseteq X$ on a  data subset $D' \subseteq D$; or a conditional PDF $\Pr_{\D'}(S | C)$ of $S$ conditioned on variables $C \subseteq X \!\setminus\! S$ on $D'$. The root node of $\mathcal{F}$ with $S = X, D' = D$ exactly represents $\Pr_{\D}(X)$. Each node $N$ decomposes the distribution $\Pr_{\D'}(S)$ or $\Pr_{\D'}(S | C)$ according to the following operations:
\vspace{-0.5em}

\hspace{1.5em} $\bullet$ \textit{Factorize}: 
Given the PDF $\Pr_{\D'}(S)$, let $H \subseteq S$ be a set of highly correlated attributes \revise{and $W = S \!\setminus\! H$}. \revise{We have $\Pr_{\D'}(S) = \Pr_{\D'}(W) \cdot \Pr_{\D'}(H | W)$}. The factorize node generates the left child and right child to process the PDF \revise{$\Pr_{\D'}(W)$} and conditional PDF \revise{$\Pr_{\D'}(H | W)$}, respectively.
\vspace{-0.5em}

\hspace{1.5em} $\bullet$ \textit{Sum}:
Given the PDF $\Pr_{\D'}(S)$, we can partition the data $D'$ into subsets $D'_1, D'_2, \dots, D'_k$. The sum node creates the $i$-th child to process the PDF $\Pr_{\D'_i}(S)$ for each $1  \! \leq \! i  \! \leq \! k$. We can regard $\Pr_{\D'}(S)$ as mixture models, i.e., $\Pr_{\D'}(S) \! = \! \! \sum_{i } w_i \Pr_{\D'_i}(S)$ where $w_i$ is the weight of the $i$-th child.
\vspace{-0.5em}

\hspace{1.5em} $\bullet$ \textit{Product}: 
Given the  PDF $\Pr_{\D'}(S)$, assume that we can partition $S$ into mutually independent subsets of variables $S_1, S_2, \dots, S_d$ on $D'$, i.e., $\Pr_{\D'}(S) = \prod_{j} \Pr_{\D'}(S_j)$. Then, the product node creates the $j$-th child to process PDF $\Pr_{\D'}(S_j)$ for each $1 \leq j \leq d$. 
\vspace{-0.5em}

\hspace{1.5em} $\bullet$ \textit{Uni-Leaf}: 
Given the PDF $\Pr_{\D'}(S)$, if $|S| = 1$, we create a leaf node with the univariate distribution $\Pr_{\D'}(S)$.
\vspace{-0.5em}

\hspace{1.5em} $\bullet$ \textit{Split}: 
\revise{Given the conditional PDF $\Pr_{\D'}(S | C)$, we partition the data $D'$ into regions $D'_1, D'_2, \dots, D'_t$ in a grid manner according to the domain space of $C$.} The $\ell$-th child models the conditional PDF $\Pr_{\D'_\ell}(S | C)$. For each value $c$ of $C$, $\Pr_{\D'}(S | c)$ is modeled on exactly one child as $\Pr_{\D'_\ell}(S | c)$. Note that the different semantic of split and sum nodes. A split node separately models $\Pr_{\D'}(S | C)$ for different values of variables in $C$ while a sum node decomposes the large model $\Pr_{\D'}(S)$ into smaller models on $S$.
\vspace{-0.5em}

\hspace{1.5em} $\bullet$ \textit{Multi-Leaf}: 
Given the conditional PDF $\Pr_{\D'}(S | C)$, if $S$ is independent of $C$ on $D'$, $\Pr_{\D'}(S | c)$ stays the same for any value $c$ of $C$. At this time, we create a leaf node to represent the multivariate distribution $\Pr_{\D'}(S)$.
   
\begin{figure*}
	\centering
	\hspace{0.05\linewidth}
	\begin{minipage}{0.35\linewidth}
	\scriptsize
	\begin{tabular}{cccc}
		\specialrule{1pt}{2pt}{2pt}
		$X_1$ {\color{mywh} $\blacksquare$}  & $X_2$ {\color{mywf} $\blacksquare$}   & $X_3$  {\color{mywt} $\blacksquare$}& $X_4$ {\color{mytp} $\blacksquare$} \\
		
		\specialrule{1pt}{2pt}{2pt}
		2 &  5.1 & 3.1 & 21\\
		4 & 8.5 & 5.6 & 1.6\\
		9 & 17.6 & 8 &  7.3 \\
		$\cdots$ & $\cdots$ & $\cdots$ & $\cdots$\\
		1 & 2.8 & 3.7 & 6.1 \\
		0 & 0.8 & 6.1 & 0.9 \\
		\specialrule{1pt}{2pt}{2pt}
	\end{tabular}
	\end{minipage}
	\hspace{0.04\linewidth}
	\begin{minipage}{0.5\linewidth}
		\includegraphics[width = 0.67\linewidth, height = 1.1in]{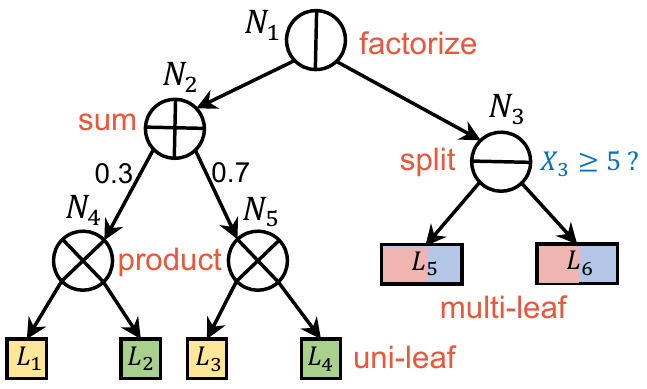}
	\end{minipage}
	\vspace{-0.5em}
	\caption{An data table of four variables and its corresponding FSPN.}
	\label{fig: fspn}
	\vspace{-2em}
\end{figure*}

The above operations are recursively applied in order to model the joint PDF. \revise{In this paper, we focus on tree-structured FSPNs. However, a more general structure for FSPNs could be a DAG. We leave the exploration of DAG-structured FSPN for future work.}
Factorize, sum, product and uni-leaf nodes represent PDFs while split and multi-leaf nodes represent conditional PDFs. Figure~\ref{fig: fspn} illustrates an FSPN example of the four variables in the data table. The two highly correlated attributes $X_1, X_2$ are first separated from $X_3, X_4$ on node $N_1$. $\Pr(X_3, X_4)$ are decomposed by sum node ($N_2$) and product nodes ($N_4, N_5$). The uni-leaf nodes $L_1, L_3$ and $L_2, L_4$ models $\Pr(X_3)$ and $\Pr(X_4)$, respectively. $\Pr(X_1, X_2 | X_3, X_4)$ are split into two regions by the value of $X_3$ on node $N_3$. 
On each region, $X_1, X_2$ are independent of $X_3, X_4$ and modeled as multi-leaf nodes $L_5$ and $L_6$.

\revise{
\textbf{Comparison with Other PGMs.}
For the remainder of this section, we compare our tree-structured FSPNs with other PGMs. First, we show that FSPNs subsume tree-structured SPNs and discrete BNs. Clearly, if we disable the factorize operation, an FSPN degenerates to the tree-structured SPN model; if we only apply the factorize, product, split, uni-leaf and multi-leaf operations, an FSPN could equally represent a discrete BN model. Due to space limits, we put the details on how a joint PDF represented by BN or tree-structured SP can be equivalently  modeled by a FSPN in Appendix~\ref{app: A}. Based on the transformation process, we obtain the following proposition, which shows that the model size of an equivalent FSPN is bounded above by the model size of a tree-structured SPN or BN, so the expressive efficiency of FSPN is no worse than them.
}

\textbf{Proposition~1}
\textit{Given a set of random variables $X$ and data matrix $D$, if $\Pr_{\D}(X)$ can be represented by an SPN or a BN with model size $M$, then there exists an equivalent FSPN
modeling $\Pr_{\D}(X)$ with model size no larger than $M$.}

\revise{
Second, we show that FSPNs resolve the drawbacks of BNs and tree-structured SPNs. FSPNs combine the strengths of both conditional and independent factorization methods. Unlike tree-structured SPNs, FSPNs separately model highly correlated variables together as multi-leaf nodes, so an FSPN's structure tends to be more compact and allows more accurate probability estimates. Unlike the situation for BNs, the structure of FSPNs can be accurately and efficiently learned, and the probability inference on an FSPN is near linear time w.r.t.~its number of nodes. We discuss the details of probability inference and structure learning for FSPNs in Section~4 and~5, respectively.}

\revise
{
Third, unlike the case of SPGMs~\citep{SPGM}, FSPNs do not require an inference process on any BN sub-structures, thus more efficient. Unlike SPN-BTBs~\citep{VarLearnSPN} and ID-SPNs~\citep{vergari2015simplifying} with multi-leaf nodes on any possible subsets $X' \subseteq X$, the multi-leaf nodes in an FSPN only model highly correlated variables $S$. Notice that values in the joint PDF of $S$ are very concentrated, so $\Pr(S)$ can be easily compressed and modeled in a lower dimension space. Whereas, for $\Pr(X')$, the storage cost for the exact PDF grows w.r.t.~$|X'|$. Hence, SPN-BTBs and ID-SPNs use BNs or MRFs with low tree-width, which may degrade the estimation accuracy and inference speed. Furthermore, DAG-SPNs~\citep{mspn,dennis2015greedy} could be subsumed by  DAG-structured FSPNs.
}

\revise{
Fourth, comparing FSPNs with CSPNs~\citep{shao2019conditional}, we note the different goals behind modeling the conditional PDFs $\Pr(Y | X)$. CSPNs try to 
 find local conditional independence between variables in $Y$ and model $\Pr(Y_i | X)$ for each $Y_i \in Y$. Whereas, FSPNs try to find local independence between $Y$ and $X$ and model $\Pr(Y) = \Pr(Y | X)$ as multi-leaf nodes together.
 Unlike SPQNs~\citep{sharir2018sum}, FSPNs use the factorize nodes to simplify the representation of joint PDF $\Pr(X, Y)$ using two simpler distributions $\Pr(X)$ and $\Pr(Y | X)$. Whereas, SPQNs model $\Pr(Y | X)$ using SPNs on $\Pr(X, Y)$ and $\Pr(X)$. 
}


\begin{figure}[!t]
	\small
	\rule{\linewidth}{1pt}
	\leftline{~~~~\textbf{Algorithm} \textsc{FSPN-Infer$(\mathcal{F}, E)$}}
	\vspace{-1em}
	\begin{algorithmic}[1]
		\STATE let $N$ be the root node of $\mathcal{F}$
		\IF{$N$ is uni-leaf node}
		\RETURN $\Pr_{\D}(E)$ by the univariate PDF on $N$
		\ELSIF{$N$ is a sum node}
		\STATE let $N_1, N_2, \dots, N_t$ be the children of $N$ with weights $w_1, w_2, \dots, w_t$
		\STATE $p_i \gets \textsc{FSPN-Infer}(\mathcal{F}_{\N_i}, E)$ for each $1 \leq i \leq t$
		\RETURN $\sum_{i = 1}^{t} w_i p_i$
		\ELSIF{$N$ is a product node}
		\STATE let $N_1, N_2, \dots, N_t$ be the children of $N$
		\STATE $p_i \gets \textsc{FSPN-Infer}(\mathcal{F}_{\N_i}, E)$ for each $1 \leq i \leq t$
		\RETURN $\prod_{i = 1}^{t} p_i$
		\ELSE
		\STATE let $N_{\L}$ be the left child on variables $W$ and $N_{\R}$ be the right child
		\STATE let $L_1, L_2, \dots, L_t$ be the multi-leaf nodes of $N_{\R}$ 
		\STATE split $E$ into $E_1, E_2, \dots, E_t$ by regions of $L_1, L_2, \dots, L_t$
		\STATE get $p_i$ of $E_i$ on variables $X \! \setminus \! W$ from the multivariate PDF on $L_i$ for each $1 \leq i \leq t$
		\STATE $p_i \gets \textsc{FSPN-Infer}(\mathcal{F}_{\N_\L}, E_i)$ for each $1 \leq i \leq t$
		\RETURN $\sum_{i = 1}^{t} p_i \cdot q_i$
		\ENDIF
	\end{algorithmic}
	\rule{\linewidth}{1pt}
	\vspace{-3em}
\end{figure}

\section{Probability Inference on FSPN}

In this section, we describe the probability inference algorithm \textsc{FSPN-Infer}. \revise{In general, \textsc{FSPN-Infer} works in a recursive manner. It starts from the root node of the FSPN and computes the probability on different types of nodes accordingly. 
Specifically, for sum or product nodes, we accumulate the probabilities from children by weighted sum or multiplication, in a way similar to tree-structured SPNs.  For each factorize node, we apply a divide-and-conquer process. Notice that, each multi-leaf node in the right child of the factorize node specifies a range, within which the highly correlated variables are locally independent of the others. Hence, we first divide the range of the computed event into several parts by  multi-leaf nodes. Then, for each part, the 
the probability of all highly correlated variables can be obtained directly from the multi-leaf node, and the probability of other variables could be recursively computed from the FSPN rooted at the left child of the factorize node. Finally, we multiply and sum them together to obtain the result.
}

Next, we formally describe the \textsc{FSPN-Infer} algorithm.
Given the FSPN $\mathcal{F}$ modeling the PDF $\Pr_{\D}(X)$, we can easily compute the marginal probability of an event of $X$. \revise{We can represent $E$ in a canonical form as a hyper-rectangle: $X_1 \in [L_1, U_1], X_2 \in [L_2, U_2], \dots, X_m \in [L_m,  U_m]$, where each side of the closed interval can also be open.} $L_i$ and $U_i$ represent the lower and upper bound of variable $X_i$, respectively. We have $L_i = -\infty$ or $U_i = \infty$ if $E$ has no constraint on left or right side of $X_i$. If the constraint of $E$ on a variable contains discontinuous intervals, we may split it into several events satisfying the above form. \revise{In this paper, we do not consider events with ranges that are not axis-aligned.}
\textsc{FSPN-Infer} takes the FSPN $\mathcal{F}$ and an event $E$ as inputs, and outputs the probability $\Pr_{\D}(E) = \Pr_{\D}(X \in E)$. 
Let $N$ be the root node of $\mathcal{F}$ (line~1). For any node $N'$ in $\mathcal{F}$, let $\mathcal{F}_{\N'}$ denote the FSPN rooted at $N'$. \textsc{FSPN-Infer} recursively computes $\Pr_{\D}(E)$ by the following rules:
\vspace{-0.5em}

\hspace{1.5em} $\bullet$ 
In the base case  (lines~2--3) where $N$ is a uni-leaf node, we directly return the probability of $E$ on the univariate PDF. 
\vspace{-0.5em}

\hspace{1.5em} $\bullet$ 
If $N$ is a sum (lines~4--7) or product node (lines~8--11), let $N_1, N_2, \dots, N_t$ be all of its children. We can further call \textsc{FSPN-Infer} on $\mathcal{F}_{\N_i}$ and $E$ for each $1 \leq i \leq t$ to obtain the probability on the PDF represented by each child. Then, node $N$ computes a weighted sum (for sum node) or multiplies (for product node) these probabilities together to obtain $\Pr_{\D}(E)$.
\vspace{-0.5em}

\hspace{1.5em} $\bullet$ 
If $N$ is a factorize node (lines~12--18), let $N_{\L}$ and $N_{\R}$ be its left child and right child. Assume $N_{\L}$ and $N_{\R}$ of $N$ representing the PDF $\Pr_{\D}(W)$ and the conditional PDF $\Pr_{\D}(H|W)$, respectively. We have $\Pr_{\D}(E) = \sum_{e \in E} \Pr_{\D}(e_{\W}) \cdot \Pr_{\D}(e_{\H} | e_{\W})$, where $e_{\W}$ and $e_{\H}$ represent the values of $e$ on variables $W$ and $H$, respectively. Let $L_1, L_2, \dots, L_t$ be all multi-leaf nodes of $N_{\R}$.
Each $L_i$ is defined on a sub-range of $W$ and maintains the PDF $\Pr_{i}(H) = \Pr_{\D}(H | w)$, which stays the same for all $w$ in this sub-range. 
Each sub-range also forms a hyper-rectangle, which is ensured by our structure learning algorithm described in the next section. For different $e_{\W}$, $\Pr_{\D}(H|e_\W)$ is represented by different PDFs on $L_i$. Therefore, we need to partition the range of $E$ into $E_1, E_2, \dots, E_t$ in terms of $W$ to compute $\Pr_{\D}(E)$. $E_i$ represents the
 sub-range of $E$ whose values of $W$ fall in $L_i$, which could also be interpreted as a valid event since its range is also a hyper-rectangle. Then, we have
\begin{equation}
\label{eq: factorize}
\small
\Pr\nolimits_{\D}(E) \! = \! \sum_{e \in E} \Pr\nolimits_{\D}(e_{\W}) \cdot \Pr\nolimits_{\D}(e_{\H} | e_{\W})
= \!\! \sum_{i=1}^{t}  \sum_{e \in E_i}  \Pr\nolimits_{\D}(e_{\H} |  e_{\W}) \cdot \Pr\nolimits_{\D}(e_{\W})
= \!\! \sum_{i=1}^{t}  \bigg[ \sum_{e_{\H} \in E_i} \Pr\nolimits_i(e_{\H}) \cdot \!\! \sum_{e_{\W} \in E_i} \Pr\nolimits_{\D}(e_{\W})\bigg].
\end{equation}
The probability $p_i \! = \! \sum_{e \in E_i} \Pr\nolimits_i(e_{\H})$ of $E_i$ on $H$ could be directly obtained from the multi-leaf $L_i$. 
The probability $q_i \! = \! \sum_{e \in E_i} \Pr\nolimits_{\D}(e_{\W})$ of $E_i$ on $W$ can be obtained by calling \textsc{FSPN-Infer} on $\mathcal{F}_{\N_{\L}}$, the FSPN rooted at the left child of $N$, and $E_i$. 
Then we sum all $p_i \cdot q_i$ together to get $\Pr_{\D}(E)$.
Similarly, for the evidence probability inference $\Pr_{\D}(Q | E \! = \! e)$ where $Q, E$ are disjoint subsets of $X$. We can
obtain $\Pr_{D}(Q, E \! = \! e)$ and $\Pr_{D}(E \! = \! e)$ on the FSPN to compute $\Pr_{\D}(Q | E \! = \! e)$.

\revise{We present a comprehensive example in Appendix~\ref{app: B}, which describes the probability inference process on an FSPN step by step. We now analyze the complexity of our inference algorithm}. Computing the probability of any range on each leaf node can be done in $O(1)$ time~\citep{LearnSPN}. Let $f$ and $l$ be the number of factorize and multi-leaf nodes in $\mathcal{F}$. 
The maximum number of ranges to be computed on each node is $O(l^f)$, so the inference time of FSPN is $O(l^f n)$.
Actually, $f$ tends to be a very small number (near $O(1)$) and the probability of multiple ranges could be computed in parallel. Therefore, we have the following proposition regarding the inference time cost of FSPN. In our evaluations,  inference on FSPN is $1$--$3$ orders of magnitude faster than BN and SPN. \revise{We reserve designing the FSPNs with theoretical bounds on $l$ and $f$ in the future work.}

\textbf{Proposition~2}
\textit{
Given a FSPN $\mathcal{F}$ representing $\Pr_{D}(X)$ with $n$ nodes and any event $E$ of $X$, the marginal probability $\Pr_{\D}(E) = \Pr_{\D}(X \in E)$ can be obtained in near $O(n)$ time on $\mathcal{F}$.}


\section{Structure Learning of FSPN}

In this section, we discuss the structure learning algorithm \textsc{Learn-FSPN} of FSPN. 
\textsc{Learn-FSPN}  takes as inputs a data matrix $D$, two sets $X$ and $C$ of variables, and outputs the FSPN for $\Pr_{\D}(X | C)$. Initially, we can call \textsc{Learn-FSPN}$(D, X, \emptyset)$ to build the FSPN on $\Pr_{\D}(X)$. \textsc{Learn-FSPN} generally works in a top-down manner by recursively identifying different operations to decompose the joint PDF into small and tractable PDFs. Due to space limits, we put the pseudocode of \textsc{Learn-FSPN} in Appendix~\ref{app: C}.
The main steps of the algorithm are described as follows:

\textit{1. Separating highly and weakly correlated variables:} 
when $C = \emptyset$, \textsc{Learn-FSPN} detects whether there exists a set $H$ of highly correlated attributes since the principle of FSPN is to separate them with others as early as possible. We find $H$ by examining pairwise correlations and then group variables whose correlation value is larger than a threshold $\tau$. If $H \neq \emptyset$, we add a factorize node to split $\Pr_{\D}(X)$.
The left child and right child recursively call \textsc{Learn-FSPN} to model $\Pr_{\D}(X \! \setminus \! H)$ and $\Pr_{\D}(H | X \! \setminus \! H)$, respectively. 

\indent \textit{2. Modeling weakly correlated variables:} 
when $C \! = \! \emptyset$ and there do not exist highly correlated variables in $X$, we try to split $\Pr_{\D}(X)$  into small regions where variables in $X$ are locally independent. 
Specifically, if $X$ contains only one variable, we model the univariate distribution $\Pr_{\D}(X)$ using maximum likelihood estimation (MLE) of parameters. This can be done by applying a multinomial Dirichlet distribution for discrete variables or a Gaussian mixture model for continuous ones. Otherwise, we first try to partition $X$ into mutually independent subsets using an independence test oracle. If $X$ can be partitioned as $S_1, S_2, \dots, S_k$, we add a product node and call \textsc{Learn-FSPN} to model $\Pr_{\D}(S_i)$ on each child.
If not, we apply an EM algorithm, \revise{such as $k$-means}, to cluster instances $D$ into $D_1, D_2, \dots, D_t$, add a sum node and call \textsc{Learn-FSPN} to model $\Pr_{\D_i}(X)$ on each child.

\indent \textit{3. Modeling conditional PDFs:} 
when $C \neq \emptyset$, it tries to model the conditional PDF $\Pr_{\D}(X | C)$. 
At this time, variables in $X$ must be highly correlated. First, we test if $X$ is independent of $C$ on $D$ by the oracle. If so, we can learn the MLE parameters of multivariate distribution $\Pr_{D}(X)$, such as the multi-dimensional Dirichlet distribution or Gaussian mixture model. 
\revise{If not, we use an EM algorithm, such as grid approximation of $k$-means, to partition the domain space of variables in $C$ into multiple grids. Based on this, instances in $D$ are split into $D'_1, D'_2, \dots, D'_t$ where each $D'_i$ represents all data points in a grid. Due to space limits, we put the details of the partition methods in Appendix~\ref{app: C}.}
Then, we add a split node and call \textsc{Learn-FSPN} to model $\Pr_{\D'_i}(X | C)$ on each child.

Note that, \textsc{Learn-FSPN} should be viewed as a framework rather than a specific algorithm, since we can choose different independent test oracles and clustering algorithms such as~\citep{neal1998view}. Moreover, any structure learning optimization~\citep{bueff2018tractable,jaini2018prometheus, kalra2018online,  BayesLearnSPN} for SPNs can be also incorporated into \textsc{Learn-FSPN}. 

Next, we show that \textsc{Learn-FSPN} returns an FSPN which locally maximizes the likelihood.
The analysis proceeds in a bottom-up manner. First, on both uni-leaf and multi-leaf nodes, the parameters of distribution are learned using MLE. Second, the independence test oracle used in the product node factorize the joint PDF into a product of independent ones, which causes no likelihood loss.
Third, for the EM methods used in sum and split nodes, the nodes can locally maximize the likelihood if all children locally maximize the likelihood~\citep{LearnSPN}. Fourth, the factorize node uses the exact probability conditional factorization, which causes no likelihood loss. Therefore, we have the following proposition. 

\textbf{Proposition~3}
\textit{
Given a set of random variables $X$ and data $D$, \textsc{Learn-FSPN} can return a local MLE FSPN $\mathcal{F}$ of $\Pr_{\D}(X)$ with independence test oracles and EM algorithms.}


\section{Evaluation Results}

In this section, we report the evaluation results on both synthetic  and real-world benchmark data.


\pgfplotsset{width = 6.4cm, height = 4.2cm}
\pgfplotsset{compat = 1.15}

\begin{figure*}	
	\centering 
	\ref{artileg}
	\begin{tikzpicture} 
	\begin{semilogyaxis}[
	xlabel = {\small Average RDC Score}, xmin = 0, xmax = 1, xtick = {0, 0.2, 0.4, 0.6, 0.8, 1.0}, xlabel shift = -4, 
	ylabel = {\small KL-divergence}, ymin = 0.00008, ymax = 0.2, ytick = {0.0001, 0.001, 0.01, 0.1}, ylabel shift = -6, ymajorgrids=true, legend columns=-1, legend style = {draw= none},
	legend to name = artileg
	]
	
	\addplot+ [ 
	mark= triangle, very thick, purple
	] coordinates {
	(0.07, 0.0001734)  
	(0.21, 0.000828)  
	(0.37, 0.00137) 
	(0.65, 0.0108)  
	(0.83, 0.01415)
	(0.95, 0.1325) 
	};
	\addlegendentry{MSPN}
	
	\addplot+ [ 
	mark= +, very thick, black
	] coordinates {
	(0.07, 0.0001734)  
	(0.21, 0.000826)  
	(0.37, 0.000917) 
	(0.65, 0.00118)  
	(0.83, 0.003156)
	(0.95, 0.004521) 
	};
	\addlegendentry{SPN-BTB}
	
	\addplot+ [ 
	mark= o, very thick, blue
	] coordinates {
	(0.07, 0.0001734)  
	(0.21, 0.000816)  
	(0.37, 0.001317) 
	(0.65, 0.00078)  
	(0.83, 0.004106)
	(0.95, 0.002621) 
	};
	\addlegendentry{SPGM}

	\addplot+ [ 
	mark= square, very thick, orange
	] coordinates {
	(0.07, 0.0001734)  
	(0.21, 0.00109)  
	(0.37, 0.000785) 
	(0.65, 0.000801)  
	(0.83, 0.00288)
	(0.95, 0.00268) 
	};
	\addlegendentry{BN-DAG}
	
	\addplot+ [ 
	mark= star, very thick, olive
	] coordinates {
	(0.07, 0.0001534)  
	(0.21, 0.00159)  
	(0.37, 0.00277) 
	(0.65, 0.00601)  
	(0.83, 0.0218)
	(0.95, 0.0762) 
	};
	\addlegendentry{BN-Tree}

	\addplot+ [ 
	mark = o, very thick, red
	] coordinates {
	(0.07, 0.0001734)  
	(0.21, 0.000833)  
	(0.37, 0.000662) 
	(0.65, 0.000607)  
	(0.83, 0.00192)
	(0.95, 0.00092) 
	};
	\addlegendentry{FSPN}

	\end{semilogyaxis}
	\end{tikzpicture}
	\hspace{2em}
	\begin{tikzpicture} 
\begin{semilogyaxis}[
xlabel = {\small{Average RDC Score}}, xmin = 0, xmax = 1, xtick = {0, 0.2, 0.4, 0.6, 0.8, 1.0}, xlabel shift = -4, 
ylabel = {\small Inference Time (ms)}, ymin = 0.3, ymax = 700, ytick = {1, 10, 100}, ylabel shift = -6, ymajorgrids=true, 
legend style={at={(0.1,0.6)},anchor=west, draw = none}, 
]

\addplot+ [ 
mark= triangle, very thick, purple
] coordinates {
	(0.07, 0.71)  
	(0.21, 10.75)  
	(0.37, 15.55) 
	(0.65, 46.5)  
	(0.83, 49.98)
	(0.95, 53.35)  
};

\addplot+ [ 
	mark= o, very thick, blue
	] coordinates {
	(0.07, 0.71)  
	(0.21, 15.15)  
	(0.37, 28.3) 
	(0.65, 76.5)  
	(0.83, 69.8)
	(0.95, 63.15) 
	};

\addplot+ [ 
mark= +, very thick, black
] coordinates {
	(0.07, 0.71)  
	(0.21, 10.35)  
	(0.37, 18.55) 
	(0.65, 66.5)  
	(0.83, 99.08)
	(0.95, 143.5)  
};

\addplot+ [ 
mark= square, very thick, orange
] coordinates {
	(0.07, 0.71)  
(0.21, 35.6)  
(0.37, 99.8) 
(0.65, 368.2)  
(0.83, 347.1)
(0.95, 373.5)  
};

\addplot+ [ 
mark= star, very thick, olive
] coordinates {
	(0.07, 25.71)  
(0.21, 45.6)  
(0.37, 49.8) 
(0.65, 86.2)  
(0.83, 107.1)
(0.95, 113.5)  
};

\addplot+ [ 
mark = o, very thick, red
] coordinates {
	(0.07, 0.71)  
(0.21, 10.81)  
(0.37, 8.25) 
(0.65, 14.1)  
(0.83, 13.59)
(0.95, 0.55)  
};
	
\end{semilogyaxis}
\end{tikzpicture}
	\vspace{-1em}
	\caption{Evaluation results on synthetic data.}
	\vspace{-1.5em}
	\label{fig: artificial}
\end{figure*}
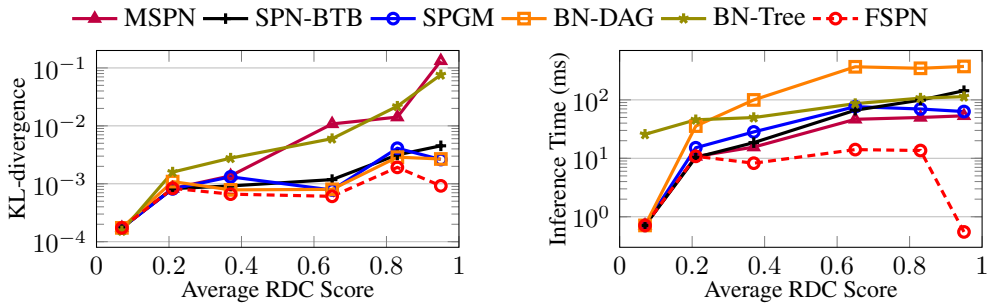

\textbf{Results on Synthetic Data.}
First, we demonstrate the superiority of FSPNs over existing PGMs on six synthetic datasets with varied degree of dependence between variables. The dependence degree is evaluated as the average RDC score~\citep{RDC} between variables, which ranges from $0.07$ to $0.95$. Each dataset contains $10^8$ rows on $20$ \revise{discrete variables with different domain size ranging from 5 to 100.}
We sampled $10^5$ rows for training PGMs and $10^4$ rows for tuning hyper-parameters. During testing, we randomly generate $10^2$ evidence queries $\Pr(Q | E = e)$. We uniformly select $2$--$4$ variables as $Q$, $6$--$10$ variables as $E$ and assign a value $e$ of $E$. To measure inference accuracy we use the KL-divergence between the true distribution and the estimated one by the learned PGM. 

\revise{
We compare our FSPN with a variety of PGMs:
MSPN is a tree-structured SPN learned using the method in~\citep{molina2018mixed} on mixed domains, which is shown to be better than the SPN learned by~\citep{LearnSPN}.
 SPN-BTB~\citep{VarLearnSPN} is a SPN structure with embedded BN as multivariate leaf nodes. The embedded BN is implemented by Chow-Liu tree whose tree width is bounded by $1$. SPGM~\citep{SPGM} is the model integrating BN and SPN. BN-DAG is a plain BN structure. BN-Tree is a BN implemented by Chow-Liu tree with bounded width of $1$. The structure of BN-DAG and BN-Tree are learned by the Pomegranate package~\citep{schreiber2018pomegranate}. 
Notice that, we do not find any open-source implementation of DAG-SPN~\citep{mspn}. Other start-of-the-art variations of BNs and SPNs, described in Table~\ref{tab: exp-binary} either work only on binary domains or do not support probability inference given evidence. Hence, we can not compare with them in our experiments. 
}

For fairness, we apply an exhaustive grid search over the hyper-parameters of each model and report the best result. Figure~\ref{fig: artificial} reports the
evaluation results in terms of the two criteria: KL-divergence for inference accuracy and the inference time for efficiency. We clearly observe that: 

\revise{
\hspace{1.5em} $\bullet$
The estimation accuracy of FSPN is consistently better than all other models. 
In comparison with MSPN and BN-Tree, the KL-divergence  decreases by up to $144 \times$ and $82\times$, respectively. This is because the 
tree-structured SPN can not work well in presence of highly correlated variables, and the tree-structured BN sacrifices the model accuracy to improve its inference speed. In comparison with SPN-BTB, SPGM and BN-DAG, FSPN also decreases the KL-divergence. This verifies that FSPN can model the joint PDF more accurately.
\vspace{-0.5em}
}

\revise{
\hspace{1.5em} $\bullet$ 
The inference time of FSPN is $1$--$3$ orders of magnitude faster than others. In comparison with BN-DAG, SPN-BTB and BN-Tree, it improves the inference speed by up to $680 \times$, $261\times$ and $206 \times$, respectively. This verifies that the BN inference process is very time costly.
Although bounding the tree width of BN could speed up the inference, it is still much slower than FSPN. FSPN is also faster than MSPN since its structure is more compact.
}

\revise{
We also evaluate the model size and training time of all models. Due to space limits, we put the results and analysis in Appendix~\ref{app: D}. 
In a nutshell, model size of learned FSPNs are up to two orders of magnitude smaller than others, and FSPNs' training time is several times faster than others except BN-Tree. In summary, this set of experiments validates the design choices of FSPNs, which provide both accurate results and fast inference.
}

\textbf{Benchmark Testing.}
Next, we compare FSPN with the current state-of-the-art methods on $20$ real-world benchmark datasets used in the literature~\citep{LearnSPN}. Table~\ref{tab: exp-binary} reports the test log-likelihood scores of FSPN and other PGMs.
Specifically, LearnSPN is the tree-structured SPN from~\citep{LearnSPN}. BayesSPN~\citep{BayesLearnSPN} is an SPN with Bayesian structure learning.  ID-SPN~\citep{VarLearnSPN} use embedded MRF as leaf nodes to enhance the performance of SPN. MT stands for the mixture of tree models~\citep{meila2000learning}.
WinMine is one of the most sophisticated BN learning package~\citep{chickering2002winmine}. \revise{
ECNet~\citep{2014Cutset} and EM-PSDD~\citep{liang2017learning} 
are the cutset network and PSDD with the best performance, respectively.
To avoid the effects of  hyper-parameters, we quote results of other PGMs from their original paper. We find that:}

\revise{
\hspace{1.5em} $\bullet$ 
Overall, FSPN outperforms LearnSPN, BayesSPN, MT, WinMine, ECNet and EM-PSDD. This is because LearnSPN and BayesSPN can not accurately model the joint PDF in presence of highly correlated variables. The expressiveness of MT is inherently low since its model complexity is not as high as others. 
\vspace{-0.5em}
}

\revise{
\hspace{1.5em} $\bullet$ 
The performance FSPN is comparable to SPGM, SPN-BTB and ID-SPN on the whole. It is slightly better than SPGM and ID-SPN but slightly worse than SPN-BTB. These models use embedded BNs or MRFs in their structure, so they are more accurate than other SPN models.}

\begin{table}
	\caption{Average test log-likelihoods on discrete datasets.}
	\medskip
	\resizebox{\columnwidth}{!}{
		\begin{tabular}{cccccccccccc}
			\specialrule{1pt}{2pt}{2pt}
			Dataset& $\#$ of vars & \textbf{FSPN (ours)} & LearnSPN & BayesSPN & SPGM & SPN-BTB & ID-SPN & MT & WinMine & ECNet & EM-PSDD \\ 
			\specialrule{1pt}{3pt}{3pt}
			NLTCS & 16 & -6.05 &-6.11 &-6.02 &\textbf{-5.99} &-6.01 &-6.00 & -6.01 &-6.03 &-6.00 &-6.03 \\
			MSNBC & 17& \textbf{-6.01} &-6.11 &-6.03 &-6.03 &-6.03 & -6.06  & -6.07 &-6.04 &-6.04 &-6.04 \\
			KDD & 65& -2.14 &-2.18 & -2.13 &-2.13 &\textbf{-2.12} & \textbf{-2.12}  & -2.13 & -2.18 &\textbf{-2.12} &\textbf{-2.12}\\
			Plants & 69& \textbf{-12.00} &-12.98 &-12.94 &-12.71 &-12.09 & -12.68  &-12.95 &-12.65 &-12.78 &-13.79 \\
			Audio & 100& -40.02 &-40.50 &-39.79 &-39.90 &\textbf{-39.62} & -39.77  &-40.08 &-40.50 &-39.73 &-41.98 \\
			Jester & 100& -52.39 &-53.48 &-52.86 &-52.83 &-53.60 & -52.42  &-53.08 &\textbf{-51.07} &-52.57 &-53.47\\
			Netflix & 100& -57.12 &-57.33 &-56.80 &-56.42 &-56.37 & -56.36  &-56.74 &-57.02 &\textbf{-56.32} &-58.41\\
			Accidents & 111& -26.99 &-30.04 &-33.89 &-26.89 &-28.35 & -26.98  &-29.63 &\textbf{-26.32} &-29.96 &-33.64\\
			Retail & 135& -10.83 & -11.04 &-10.83 &-10.83 &-10.86 & -10.88  &-10.83 &-10.87 &-10.82 & \textbf{-10.81}\\
			Pumsb-star & 163& \textbf{-22.04} &-24.78 & -31.96 &-22.15 &-22.66 & -22.40  &-23.71 &-22.72 &-24.18 &-33.67\\
			DNA & 180& -80.97 &-82.52 &-92.84 &\textbf{-79.88} &-80.07 & -81.21  &-85.14 &-80.65 &-85.82 &-92.67\\
			Kosarak & 190& -10.66 &-10.99 &-10.77 &\textbf{-10.57} &-10.58 & -10.60  &-10.62 &-10.83 &-10.58 &-10.81\\
			MSWeb & 294&\textbf{-9.60} & -10.25 &-9.89 &-9.81 &-9.61 & -9.73  &-9.85 &  -9.70 & -9.79 &-9.97\\
			Book & 500& \textbf{-33.81} &-35.89 &-34.34 &-34.18 &-33.82 & -34.14  &-34.63 &-36.41 &-33.96 &-34.97\\
			EachMovie & 500& -50.69 &-52.49 &-50.94 &-54.08 &\textbf{-50.41} & -51.51  &-54.60 &-54.37 &-51.39 &-58.01\\
			WebKB & 839& \textbf{-149.72} &-158.20 &-157.33 &-154.55 &-149.85 & -151.84  &-156.86 &-157.43 &-153.22 &-161.09\\
			Reuters-52 & 889& -81.62 &-85.07 &-84.44 &-85.24 &\textbf{-81.59} & -83.35  &-85.90 &-87.55 &-86.11 &-89.61\\
			20 Newsgrp & 910& -155.30 &-155.93 &-151.95  &-153.69 &--- & -151.47 &-154.24 &-158.95 &\textbf{-151.29} &-161.09\\
			BBC & 1, 058& -252.81 &-250.69 &-254.69 &-255.22 &\textbf{-226.56} & -248.93  &-261.84 &-257.86 &-250.58 &-253.19 \\
			AD & 1, 556& -15.46 & -19.73 &-63.80 &-14.30 &\textbf{-13.60} & -19.00  &-16.02 &-18.35 &-16.68 &-31.78\\
			\specialrule{1pt}{2pt}{2pt}
	\end{tabular}}
	\label{tab: exp-binary}
	\vspace{-1em}
\end{table}

\section{Conclusions}

In this paper we propose FSPNs, \revise{a novel class of PGMs aiming at overcoming the drawbacks of existing PGMs. FSPN can adaptively model the joint distribution of variables with different dependence degree. It achieves high estimation accuracy and tractability at the same time.} We design a near linear time marginal probability inference algorithm and a local MLE structure learning algorithm for FSPNs. \revise{Our extensive evaluation results on synthetic and benchmark datasets demonstrate that FSPNs attain superior performance.} Based on these promising results, we
affirmatively believe in that FSPNs \revise{may be a better alternative to} existing PGMs in a wide range of ML applications. Moreover, \revise{we believe that there are many possible extensions of} FSPNs worth researching in the future, such as supporting maximum a posterior inference, latent variable interpretation, \revise{DAG and Bayesian structure leaning of FSPNs and bounding their tree-width for tractability}.

\clearpage

\bibliography{iclr2021_conference}
\bibliographystyle{iclr2021_conference}

\appendix


\clearpage

\section{Generality of FSPN}
\label{app: A}

We present the details on how FSPNs subsume \revise{tree-structured} SPNs, as well as discrete BNs, where all variables are discretized and all (conditional) probability distributions are stored in a tabular form.

\hspace{1.5em} $\bullet$ 
Given a set of variables $X$ and data $D$, if $\Pr_{D}(X)$ could be represented by \revise{a tree-structured} SPN $\mathcal{S}$, we can easily construct an FSPN $\mathcal{F}$ that equally represent $\Pr_{D}(X)$. Specifically, we disable the factorize operation in FSPN by setting the factorization threshold to $\infty$, and follow the same steps of $\mathcal{S}$ to construct $\mathcal{F}$. Then, the FSPN $\mathcal{F}$ is exactly the same of $\mathcal{S}$. Obviously, their model size is the same.

\begin{figure}[h]
	\small
	\rule{\linewidth}{1pt}
	\leftline{~~~~\textbf{Algorithm} \textsc{BN-to-FSPN$(\mathcal{B}, N)$}}
	\vspace{-1em}
	\begin{algorithmic}[1]
		\IF{$\mathcal{B}$ contains more than one connected component $\mathcal{B}_1, \mathcal{B}_2, \dots, \mathcal{B}_t$}
			\STATE set $N$ to be a product node with children $N_1, N_2, \dots, N_t$
			\STATE call \textsc{BN-to-FSPN$(\mathcal{B}_i, N_i)$}
			for each $i$
		\ELSE
			\STATE let $X_i$ be a node in $\mathcal{B}$ containing no out-neighbor
			\IF{$X_i$ has no in-neighbor in $\mathcal{B}$ }
				\STATE set $N$ to be a uni-leaf node representing $\Pr_{D}(X_i)$
			\ELSE
				\STATE  set $N$ to be a factorize node with left child $N_\L$ and right child $N_\R$
				\STATE set $N_\R$ to be a split node
				\FOR{each value $y$ of $X_{\pa(i)}$ in the CPT of $X_i$}
					\STATE add a multi-leaf node $N_y$ as child of $N_\R$
					\STATE let $D_y \gets \{ d \in D | X_{\pa(i)} \text{ of } d \text{ is } y\}$ 
					\STATE let $N_y$ represent $\Pr_{D_y}(X_i)$
				\ENDFOR
				\STATE remove $X_i$ from $\mathcal{B}$ to be $\mathcal{B'}$
				\STATE call \textsc{BN-to-FSPN$(\mathcal{B}', N_\L)$}
			\ENDIF
		\ENDIF
	\end{algorithmic}
	\rule{\linewidth}{1pt}
	\vspace{-2em}
\end{figure}

\hspace{1.5em} $\bullet$ 
Given a set of variables $X$ and data $D$, if $\Pr_{\D}(X)$ can be represented by a discrete BN $\mathcal{B}$, we can also build an FSPN $\mathcal{F}$ that equally represent $\Pr_{\D}(X)$. Without ambiguity, we also use $\mathcal{B}$ to refer to its DAG structure.
We present the procedures in the \textsc{BN-to-FSPN} algorithm.
It takes as inputs a discrete BN $\mathcal{B}$ and a node $N$ in $\mathcal{F}$ and outputs $F_\N$ representing the PDF of $\mathcal{B}$. We initialize $\mathcal{F}$ with a root node $N$.
Then, \textsc{BN-to-FSPN} works in a recursive manner by executing the following steps:

\hspace{1.5em} \textcircled{1} \underline{(lines~1--3)}
If $\mathcal{B}$ contains more than one connected component
$\mathcal{B}_1, \mathcal{B}_2, \dots, \mathcal{B}_t$, the variables in each are mutually independent. Therefore, we set $N$ to be a product node with children $N_1, N_2, \dots, N_t$ into $\mathcal{F}$, and call \textsc{BN-to-FSPN} on $\mathcal{B}_i$ and node $N_i$ for each $i$.

\hspace{1.5em} \textcircled{2} \underline{(lines~5--7)}
If $\mathcal{B}$ contains only one connected component, let $X_i$ be a node (variable) in $\mathcal{B}$ that has no out-neighbor. If $X_i$ also has no in-neighbor (parent) in $\mathcal{B}$, it maintains the PDF $\Pr_{\D}(X_i)$. At that time, we set $N$ to be a uni-leaf representing the univariate distribution $\Pr_{\D}(X_i)$. 

\hspace{1.5em} \textcircled{3} \underline{(lines~9--16)}
If the parent set $X_{\pa(i)}$ of $X_i$ is not empty, $X_i$ has a conditional probability table (CPT) defining $\Pr_{\D}(X_i | X_{\pa(i)}) = \Pr_{\D}(X_i | X \setminus \{X_i\})$. At this time, 
we set $N$ to be a factorize node with the left child representing $\Pr_{\D}(X \setminus \{X_i\})$ and right child $N_{\R}$ representing $\Pr_{\D}(X_i | X \setminus \{X_i\})$. For the right child $N_{\R}$, we set it to be a split node. For each entry $y$ of $X_{\pa(i)}$ in the CPT of $X_i$, we add a leaf $L_y$ of $N_{\R}$ containing all data $D_y$ in $D$ whose value on $X_{\pa(i)}$ equals $y$. On each leaf $L_y$, by the first-order Markov property of BN, $X_i$ is conditionally independent of variables $X \setminus \{ X_i \} \setminus X_{\pa(i)}$ given its parents $X_{\pa(i)}$. Therefore, we can simplify the PDF represented by $L_y$ as $\Pr_{\D}(X_i | y) = \Pr_{\D_y}(X_i)$.
Therefore, $N_{\R}$ characterizes the CPT of $\Pr_{\D}(X_i | X_{\pa(i)}) = \Pr_{\D}(X_i | X \setminus \{X_i\})$. 
\vspace{-0.5em}

\hspace{1.5em} Later, we remove the node $X_i$ from $\mathcal{B}$ to be $\mathcal{B'}$, which represents the PDF $\Pr_{\D}(X \setminus \{X_i\})$. We call \textsc{BN-to-FSPN} on $\mathcal{B'}$ and node $N_{\L}$, the left child of $N$ to further model the PDF.

Finally, we obtain the FSPN $\mathcal{F}$ representing the same PDF of $\mathcal{B}$. Next, we analyze the model size of $\mathcal{B}$ and $\mathcal{F}$. The storage cost of each node $X_i$ in $\mathcal{B}$ is the number of entries in CPT of $\Pr_{\D}(X_i | X_{\pa(i)})$. The FSPN $\mathcal{F}$ represents $\Pr_{\D}(X_i)$ in step \textcircled{2} when $X_{\pa(i)}$ is empty and $\Pr_{\D}(X_i | y)$ for each value $y$ of $X_{\pa(i)}$ in step \textcircled{3}. In the simplest case, if $\mathcal{F}$ also represents the distribution in a tabular form, the storage cost is the same as $\mathcal{B}$. Therefore, the model size of $\mathcal{F}$ can not be larger than that of $\mathcal{B}$. 

Consequently, proposition~1 holds.

\section{Example of Probability Inference on FSPN}
\label{app: B}

\revise{
We show an example of probability inference on the FSPN in Figure~\ref{fig: fspn}. This FSPN models the joint PDF on four variables $X_1, X_2, X_3, X_4$. The two highly correlated attributes $X_1, X_2$ are first separated from $X_3, X_4$ on node $N_1$. $\Pr(X_3, X_4)$ are decomposed by sum node ($N_2$) and product nodes ($N_4, N_5$). The uni-leaf nodes $L_1, L_3$ and $L_2, L_4$ models $\Pr(X_3)$ and $\Pr(X_4)$, respectively. $\Pr(X_1, X_2 | X_3, X_4)$ are split into two regions by the value of $X_3$ on node $N_3$. 
On each region, $X_1, X_2$ are independent of $X_3, X_4$ and modeled as multi-leaf nodes $L_5$ and $L_6$.}

\begin{figure*}
	\centering
	\begin{minipage}{0.5\linewidth}
		\includegraphics[width = 0.67\linewidth, height = 1.1in]{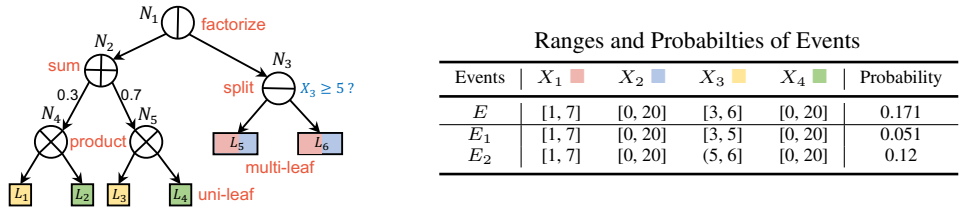}
	\end{minipage}
	\hspace{-0.1\linewidth}
	\begin{minipage}{0.4\linewidth}
		\small
		\text{~~~~~~~~~~~~~~~~Ranges and Probabilties of Events}
		\scriptsize
		\\
		\begin{tabular}{c|cccc|c}
			\specialrule{1pt}{2pt}{2pt}
			Events & $X_1$ {\color{mywh} $\blacksquare$}  & $X_2$ {\color{mywf} $\blacksquare$}   & $X_3$  {\color{mywt} $\blacksquare$}& $X_4$ {\color{mytp} $\blacksquare$} & Probability \\
			\specialrule{1pt}{2pt}{2pt}
			$E$ & [1, 7] &  [0, 20] & [3, 6] & [0, 20] &0.171 \\ \hline
			$E_1$ & [1, 7] &  [0, 20] & [3, 5] & [0, 20] & 0.051\\
			$E_2$ & [1, 7] &  [0, 20] & (5, 6] & [0, 20] & 0.12\\
			\specialrule{1pt}{2pt}{2pt}
		\end{tabular}
	\end{minipage}
	\vspace{-0.5em}
	\caption{An example of probability inference on  FSPN.}
	\label{fig: fspn-app}
	\vspace{-0.5em}
\end{figure*}

\revise{
We assume that the domain of each variable is $[0, 20]$. 
We consider the example event $E: X_1 \in [1, 7], X_3 \in [3, 6]$, whose canonical form is $E: X_1 \in [1, 7], X_2 \in [0, 20], X_3 \in [3, 6], X_4 \in [0, 20]$. We then obtain its probability on the FSPN by following the procedures step by step in Figure~\ref{fig: fspn-app}.
}

\revise{
First, we consider the factorize root node $N_1$. The right child $N_3$ splits the domain on whether $X_3$ is greater than $5$ to the two multi-leaf nodes $L_5$ and $L_6$. The domains on $L_5$ and $L_6$ could be represented as $X_1 \in [0, 20], X_2 \in [0, 20], X_3 \in [0, 5], X_4 \in [0, 20]$ and $X_1 \in [0, 20], X_2 \in [0, 20], X_3 \in (5, 20], X_4 \in [0, 20]$, respectively. We divide the range of $E$ into two parts by intersecting with the domains of $L_5$ and $L_6$ as $E_1: X_1 \in [1, 7], X_2 \in [0, 20], X_3 \in [3, 5], X_4 \in [0, 20]$ and $E_2: X_1 \in [1, 7], X_2 \in [0, 20], X_3 \in (5, 6], X_4 \in [0, 20]$. Obviously, we have $\Pr(E) = \Pr(E_1) + \Pr(E_2)$.
}

\revise{
Second, we consider how to compute $\Pr(E_1)$ and $\Pr(E_2)$.
Within the range of $E_1$, the variables $X_3, X_4$ is locally independent of the highly correlated variables $X_1, X_2$, so we have $\Pr(E_1) = \Pr(X_1, X_2) \cdot \Pr(X_3, X_4)$. The joint PDF of  highly correlated variables $X_1, X_2$ is modeled together by the multivariate leaf node $L_5$, so we get the probability $\Pr(X_1 \in [1, 7], X_2 \in [0, 20]) = 0.3$ from $L_5$. The joint PDF of variables $X_3, X_4$ is modeled by the FSPN rooted at node $N_2$, the left child of $N_1$. We can obtain the probability in a similar way of SPN. Specifically, $N_2$ is a sum node, so we have  $\Pr_{\N_2}(X_3 \in [3, 5], X_4 \in [0, 20]) = 0.3 \cdot \Pr_{\N_4}( X_3 \in [3, 5], X_4 \in [0, 20] ) + 0.7 \cdot \Pr_{\N_5}( X_3 \in [3, 5], X_4 \in [0, 20] ) $. $N_4$ is a product node, so we have $\Pr_{\N_4}( X_3 \in [3, 5], X_4 \in [0, 20] )  = \Pr_{\L_1}( X_3 \in [3, 5]) \cdot \Pr_{\L_2}( X_4 \in [0, 20])$. The probability that $\Pr_{\L_1}( X_3 \in [3, 5]) = 0.1$ and $\Pr_{\L_2}(X_4 \in [0, 20]) = 1$ could be obtained from the univariate leaf nodes $L_1$ and $L_2$, respectively. Hence, we get $\Pr_{\N_4}( X_3 \in [3, 5], X_4 \in [0, 20] )  = 0.1$. The probability $\Pr_{\N_5}( X_3 \in [3, 5], X_4 \in [0, 20] ) = 0.2$ could be obtained from leaf nodes $L_3$ and $L_4$ in the same way. As a result, we have $\Pr_{\N_2}(X_3 \in [3, 5], X_4 \in [0, 20]) = 0.3 * 0.1 + 0.7 * 0.2 = 0.17$ and $\Pr(E_1) = 0.3 * 0.17 = 0.051$.
}

\revise{
Third, the probability of $E_2$ could be computed in the same way as $E_1$. For $E_2$, the probability $\Pr(X_1 \in [1, 7], X_2 \in [0, 20]) = 0.4$ is obtained from the multivariate leaf node $L_6$, and the probability $\Pr_{\N_2}(X_3 \in (5, 6], X_4 \in [0, 20]) = 0.3$ is also computed by the FSPN rooted at node $N_2$. We have $\Pr(E_2) = 0.4 * 0.3 = 0.12$.
}

\revise{
Finally, we obtain the probability of $E$ as $\Pr(E) = \Pr(E_1) + \Pr(E_2) = 0.171$.
}

\section{Details of the \textsc{Learn-FSPN} Algorithm}
\label{app: C}

\begin{figure}[h]
	\small
	\rule{\linewidth}{1pt}
	\leftline{~~~~\textbf{Algorithm} \textsc{Learn-FSPN$(D, X, C)$}}
	\vspace{-1em}
	\begin{algorithmic}[1]
		\IF{$C = \emptyset$}
			\STATE test the correlations $c_{ij}$ for each pair of attributes $X_i, X_j \in X$
			\STATE $H \gets \{X_i , X_j | c_{ij} \geq \tau \}$
			\STATE recursively enlarge $H \gets H \cup \{X_k |  c_{ik} \geq \tau, X_i \in H, X_k \in X \setminus H \}$
			\IF{$H \neq \emptyset$}
			\STATE set $N$ to be a factorize node
			\STATE call $\textsc{Learn-FSPN}(D, X \setminus H, \emptyset)$ 
			\STATE call $\textsc{Learn-FSPN}(D, H, X \setminus H)$ 
			\ELSIF{$|X| = 1$}
			\STATE set $N$ to be a uni-leaf node
			\STATE model the univariate distribution $\Pr_{\D}(X)$ 
			\ELSIF{subsets $S_1, S_2, \dots, S_k$ are mutually indepedent by the independence oracle}
			\STATE set $N$ to be a product node
			\STATE call $\textsc{Learn-FSPN}(D, S_i, \emptyset)$ for each $1 \leq i \leq k$
			\ELSE
			\STATE set $N$ to be a sum node
			\STATE let $D_1, D_2, \dots, D_t$ be the generated cluster of data with weights $w_1, w_2, \dots, w_t$
			\STATE $w_i \gets {|T_i|}/{|T_\N|}$ for all $1 \leq i \leq n$
			\STATE call $\textsc{Learn-FSPN}(D_i, X, \emptyset)$ for each $1 \leq i \leq t$
		\ENDIF
	\ELSE
		\IF{$X$ is independent of $C$ on $D$ by the independence oracle}
			\STATE set $N$ to be a multi-leaf node
			\STATE model the  multivariate distribution $\Pr_{D}(X)$
		\ELSE
			\STATE set $N$ to be a split node
			\STATE let $D'_1, D'_2, \dots, D'_t$ be the generated partition of data
			\STATE call $\textsc{Learn-FSPN}(D'_i, X, C)$ for each $1 \leq i \leq t$
		\ENDIF
	\ENDIF
	\end{algorithmic}
	\rule{\linewidth}{1pt}
\end{figure}

\revise{
In our implementation of the \textsc{Learn-FSPN} algorithm, we use the RDC score~\citep{RDC} as the independence test oracle since it can capture dependencies between variables of hybrid domains. Two variables are identified to be independent and highly correlated if their RDC score is lower than a threshold $\tau_{\L}$ or larger than a threshold $\tau_{\H}$, respectively.
In our experiment, we set $\tau_{\L} = 0.3$ and $\tau_{\H} = 0.7$, respectively. 
}

\revise{For the clustering method in line~17 for sum nodes, we use $k$-means, an EM method. For the partition method in line~26 for split nodes, we design two methods as follows:
}

\revise{
\hspace{1.5em} $\bullet$ 
Grid approximation of $k$-means: 
At first, we use the $k$-means method to cluster all data into two clusters. By the properties of $k$-means, each clustering forms a hyper-spheroid in the space. Let $c_1$ and $c_2$ be the central points and $r_1$ and $r_2$ by the radius of the two clusters, respectively. Certainly, on the straight line across $c_1$ and $c_2$ in the space, there must exist two boundary points $b_1$ and $b_2$ of the two clusters. Let $b$ be the mid-point of $b_1$ and $b_2$, we can split all data into two parts by one dimension of $b$'s value. Some data points $x$ in one cluster would be divided into the other part. Hence, we choose the dimension of $b$'s value with minimal $|x|$ as the splitting point.
}

\revise{
\hspace{1.5em} $\bullet$ 
Greedy splitting: Let $c \in C$ be the variable that maximizes the pairwise correlations between variables in $X \setminus C$ and $C$. Intuitively, dividing the space by $c$ would largely break the correlations between $X \setminus C$ and $C$. Then, we randomly choose $d$ values $c_1, c_2, \dots, c_d$ in the range of variable $c$. For each value $c_i$, we could divide all data into two parts. We compute the pairwise correlations between variables in $X \setminus C$ and $C$ in each part. The value $c_i$ minimizes this value is chosen as the splitting point. 
}

\revise{
Notice that, \textsc{Learn-FSPN} is a framework support different independent test oracles, clustering and partition algorithms. The problem to design best plug-ins of these methods for a specific application is still open. In our experiments, we have also tried some other kinds of independent test and clustering methods. We find that using RDC scores, $k$-means clustering and greedy splitting methods attains the best performance on our datasets. 
}

\section{Additional Experimental Results}
\label{app: D}

\revise{
We present additional experimental results on the model size and training time of PGMs. Figure~\ref{fig: artificial-app} shows the model size and training time of each PGM on the synthetic datasets. Table~\ref{fig: modelsize} gives the detailed number of nodes in each model.}

\revise{In terms of the model and number of nodes, we observe that:}

\revise{
\hspace{1.5em} $\bullet$
The model and number of nodes of FSPNs are consistently much smaller than SPN-BTBs and MSPNs  by up to two orders of magnitude. In particular, the  FSPNs' model size is up to $27 \times$ and $116\times$ smaller than MSPNs and SPN-BTBs, respectively. The FSPNs' number of nodes is up to $126 \times$ and $73\times$ smaller than MSPNs and SPN-BTBs, respectively.
This is because tree-structured SPNs may generate a large number of nodes in presence of highly correlated variables. SPN-BTBs have lots of leaf nodes, and for each leaf node, they create a Chow-Liu tree. As a result, the space cost of SPN-BTBs is the highest among all models. 
}

\revise{
\hspace{1.5em} $\bullet$
The model size of FSPNs is also smaller than SPGMs and BN-DAGs. This is because they require to store the CPTs over multiple variables, while FSPNs store the lightweight univariate distributions and multivariate distributions on only highly correlated variables.  In terms of the number of nodes, FSPNs are also much smaller than SPGMs by up to $17\times$. The number of nodes in BN-DAGs and BN-Trees always equals to the variable number so it is not informative to make a comparison. 
}

\revise{
\hspace{1.5em} $\bullet$
The model size of BN-Tree is the smallest since the tree-width is only one for the learned Chow-Liu tree.
}

\revise{
In terms of the training time, we find that FSPNs are several times faster to train than other models, except BN-Trees. Whereas for other models (BN-DAGs, SPN-BTBs, SPGMs) related to BNs, the structure learning process is time costly. MSPNs also require a
longer training time since the structure learning algorithm repeatedly splits nodes in presence of highly correlated variables. 
}

\revise{
	In summary, these results verify our claims in Section~2. The results show that the design choices underlying FSPNs,  i.e. separating highly correlated variables from others and modeling them adaptively may represent the joint PDF in a more compact form and learn it efficiently.
}


\pgfplotsset{width = 6.4cm, height = 4.2cm}
\pgfplotsset{compat = 1.15}

\begin{figure*}[t]
	\centering
	\hspace{15em} \ref{artileg-app}
\begin{tikzpicture} 
\begin{semilogyaxis}[
xlabel = {Average RDC Score}, xmin = -0.01, xmax = 1.01, xtick = {0, 0.2, 0.4, 0.6, 0.8, 1.0}, xlabel shift = -4, 
ylabel = {Model Size (KB)}, ymin = 2, ymax = 2000, ytick = {10, 100}, ylabel shift = -6, ymajorgrids=true, 	legend columns=-1, legend style = {draw= none},
legend to name = artileg-app
]

\addplot+ [ 
mark= triangle, very thick, purple
] coordinates {
	(0.07, 5.34)  
	(0.21, 25.1)  
	(0.37, 99.3) 
	(0.65, 375.8)  
	(0.83, 403.6)
	(0.95, 362.2)   
};
\addlegendentry{MSPN}

\addplot+ [ 
mark= +, very thick, black
] coordinates {
	(0.07, 5.34)  
	(0.21, 24.2)  
	(0.37, 169.7) 
	(0.65, 375.8)  
	(0.83, 1713.6)
	(0.95, 1562.2)   
};
\addlegendentry{SPN-BTB}

\addplot+ [ 
mark= o, very thick, blue
] coordinates {
	(0.07, 5.34)  
	(0.21, 34.55)  
	(0.37, 55.7) 
	(0.65, 72.83)  
	(0.83, 210.61)
	(0.95, 264.1)   
};
\addlegendentry{SPGM}

\addplot+ [ 
mark= square, very thick, orange
] coordinates {
	(0.07, 5.34)  
	(0.21, 30.1)  
	(0.37, 58.5) 
	(0.65, 84.7)  
	(0.83, 116.1)
	(0.95, 154.9)  
};
\addlegendentry{BN-DAG}

\addplot+ [ 
mark= star, very thick, olive
] coordinates {
	(0.07, 5.34)  
	(0.21, 28.1)  
	(0.37, 35.5) 
	(0.65, 34.7)  
	(0.83, 56.1)
	(0.95, 44.9)  
};
\addlegendentry{BN-Tree}

\addplot+ [ 
mark = o, very thick, red
] coordinates {
	(0.07, 5.34)  
	(0.21, 25.9)  
	(0.37, 31.6) 
	(0.65, 67.9)  
	(0.83, 102.5)
	(0.95, 13.4)  
};
\addlegendentry{FSPN}

\end{semilogyaxis}
\end{tikzpicture}
\hspace{2em}
\begin{tikzpicture} 
\begin{semilogyaxis}[
xlabel = {Average RDC Score}, xmin = 0, xmax = 1, xtick = {0, 0.2, 0.4, 0.6, 0.8, 1.0}, xlabel shift = -4, 
ylabel = {Training Time (min)}, ymin = 0.3, ymax = 100, ytick = {1, 10, 100}, ylabel shift = -6, ymajorgrids=true, 
]

\addplot+ [ 
mark= triangle, very thick, purple
] coordinates {
	(0.07, 0.55)  
	(0.21, 10.3)  
	(0.37, 16.72) 
	(0.65, 31.02)  
	(0.83, 33.98)
	(0.95, 33.79)  
};

\addplot+ [ 
	mark= o, very thick, blue
	] coordinates {
	(0.07, 0.55)  
	(0.21, 15.15)  
	(0.37, 18.13) 
	(0.65, 28.5)  
	(0.83, 27.18)
	(0.95, 33.15) 
	};

\addplot+ [ 
mark= +, very thick, black
] coordinates {
	(0.07, 0.55)  
	(0.21, 10.83)  
	(0.37, 12.52) 
	(0.65, 23.61)  
	(0.83, 27.08)
	(0.95, 33.14)  
};

\addplot+ [ 
mark= square, very thick, orange
] coordinates {
	(0.07, 1.35)  
(0.21, 12.21)  
(0.37, 37.18) 
(0.65, 58.26)  
(0.83, 67.12)
(0.95, 53.52)  
};

\addplot+ [ 
mark= star, very thick, olive
] coordinates {
	(0.07, 2.11)  
(0.21, 2.61)  
(0.37, 2.83) 
(0.65, 3.12)  
(0.83, 4.13)
(0.95, 3.51)  
};

\addplot+ [ 
mark = o, very thick, red
] coordinates {
	(0.07, 0.55)  
(0.21, 5.87)  
(0.37, 8.23) 
(0.65, 13.12)  
(0.83, 15.59)
(0.95, 2.1)  
};

\end{semilogyaxis}
\end{tikzpicture}

	\vspace{-1em}
	\caption{Model size and training time on synthetic data.}
	\vspace{-1em}
	\label{fig: artificial-app}
\end{figure*}
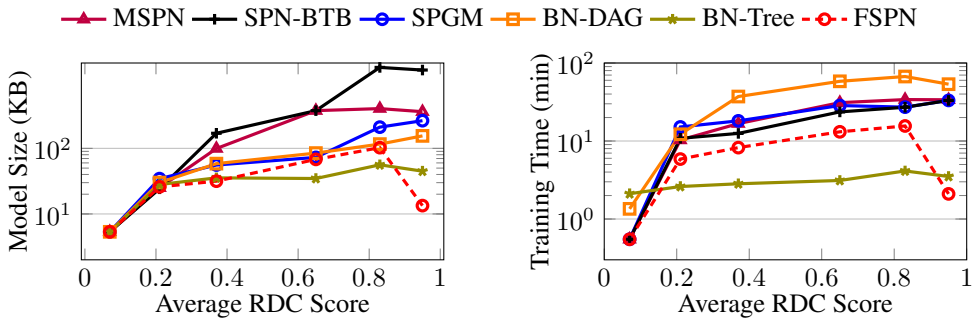

\begin{table*}[t]
\caption{Number of nodes on synthetic data.}
\resizebox{\linewidth}{!}{
\begin{tabular}{ccccccc}
	\specialrule{1pt}{2pt}{2pt}
   RDC Score& $\#$ of nodes & $\#$ of nodes & $\#$ of nodes & $\#$ of nodes & $\#$ of nodes \\
	of Dataset & in MSPN & in SPN-BTB & in SPGM & in BN-DAG/Tree & in FSPN \\
	\specialrule{1pt}{2pt}{2pt}
	0.07 &  21 &21 &21 & 20 & 21\\
	0.21 & 765 & 562 & 321 & 20 & 168\\
	0.37 & 1, 482 & 1, 175 & 472 & 20 & 375 \\
	0.65 & 3, 520 & 1, 626 & 481 & 20 & 437\\
	0.83 & 3, 847 & 2, 753 & 718 & 20 & 661\\
	0.95 & 4, 173 & 2, 398 & 563 & 20 & 33 \\
	\specialrule{1pt}{2pt}{2pt}
\end{tabular}}
\vspace{2em}
\label{fig: modelsize}
\end{table*}

\end{document}